\documentclass[a4paper,11pt]{article}
\usepackage{float}
\usepackage{tabularx,multirow}
\usepackage{rotating}
\usepackage{graphicx}
\usepackage[utf8]{inputenc}
\usepackage{acl2013}
\usepackage{times}
\usepackage{url}
\usepackage{latexsym}
\usepackage{amsmath}
\usepackage{array}
\newcolumntype{L}[1]{>{\raggedright\arraybackslash}p{#1}}
\newcolumntype{C}[1]{>{\centering\arraybackslash}m{#1}}

\title{The CoNLL-2013 Shared Task on Grammatical Error Correction}
\author{Hwee Tou Ng \\
  Department of Computer Science \\
  National University of Singapore \\
  {\tt nght@comp.nus.edu.sg} \\\And
  Siew Mei Wu \\
  Centre for English Language Communication \\
  National University of Singapore \\
  {\tt elcwusm@nus.edu.sg} \\\AND
  Yuanbin Wu \and Christian Hadiwinoto \\
  Department of Computer Science \\
  National University of Singapore \\
  {\tt \{wuyb,chrhad\}@comp.nus.edu.sg} \\\And
  Joel Tetreault \\
  Nuance Communications, Inc. \\
  {\tt Joel.Tetreault@nuance.com}
}

\date{}

\begin{document}
\maketitle
\begin{abstract}

The CoNLL-2013 shared task was devoted to grammatical error
correction.  In this paper, we give the task definition, present the
data sets, and describe the evaluation metric and scorer used in the
shared task. We also give an overview of the various approaches
adopted by the participating teams, and present the evaluation
results.

\end{abstract}

\section{Introduction}
\label{sec:introduction}

Grammatical error correction is the shared task of the Seventeenth
Conference on Computational Natural Language Learning in 2013
(CoNLL-2013). In this task, given an English essay written by a
learner of English as a second language, the goal is to detect and
correct the grammatical errors present in the essay, and return the
corrected essay.

This task has attracted much recent research interest, with two shared
tasks Helping Our Own (HOO) 2011 and 2012 organized in the past two
years \cite{dale_hoo_2011,dale_hoo_2012}. In contrast to previous
CoNLL shared tasks which focused on particular subtasks of natural
language processing, such as named entity recognition, semantic role
labeling, dependency parsing, or coreference resolution, grammatical
error correction aims at building a complete end-to-end
application. This task is challenging since for many error types,
current grammatical error correction systems do not achieve high
performance and much research is still needed. Also, tackling this
task has far-reaching impact, since it is estimated that hundreds of
millions of people worldwide are learning English and they benefit
directly from an automated grammar checker.

The CoNLL-2013 shared task provides a forum for participating teams to
work on the same grammatical error correction task, with evaluation on
the same blind test set using the same evaluation metric and
scorer. This overview paper contains a detailed description of the
shared task, and is organized as follows.  Section~\ref{sec:task}
provides the task definition. Section~\ref{sec:data} describes the
annotated training data provided and the blind test
data. Section~\ref{sec:eval} describes the evaluation metric and the
scorer. Section~\ref{sec:approaches} lists the participating teams
and outlines the approaches to grammatical error correction used by
the teams.  Section~\ref{sec:results} presents the results of the
shared task. Section~\ref{sec:conclusions} concludes the paper.

\section{Task Definition}
\label{sec:task}

The goal of the CoNLL-2013 shared task is to evaluate algorithms and
systems for automatically detecting and correcting grammatical errors
present in English essays written by second language learners of
English. Each participating team is given training data manually
annotated with corrections of grammatical errors. The test data
consists of new, blind test essays. Preprocessed test essays, which
have been sentence-segmented and tokenized, are also made available to
the participating teams. Each team is to submit its system output
consisting of the automatically corrected essays, in
sentence-segmented and tokenized form.

Grammatical errors consist of many different types, including articles
or determiners, prepositions, noun form, verb form, subject-verb
agreement, pronouns, word choice, sentence structure, punctuation,
capitalization, etc. Of all the error types, determiners and
prepositions are among the most frequent errors made by learners of
English. Not surprisingly, much published research on grammatical
error correction focuses on article and preposition errors
\cite{han06,gamon10,rozovskaya10,tetreault10,dahlmeier_acl11}, with
relatively less work on correcting word choice errors
\cite{dahlmeier_emnlp11}.  Article and preposition errors were also
the only error types featured in the HOO 2012 shared task. Likewise,
although all error types were included in the HOO 2011 shared task,
almost all participating teams dealt with article and preposition
errors only (besides spelling and punctuation errors).

In the CoNLL-2013 shared task, it was felt that the community is now
ready to deal with more error types, including noun number, verb form,
and subject-verb agreement, besides articles/determiners and
prepositions. Table~\ref{tab:error-types} shows examples of the five
error types in our shared task.

\begin{table*}
\begin{center}
\begin{tabularx}{\textwidth}{|l|l|X|l|} \hline
\multicolumn{1}{|c|}{\bf Error tag} & \multicolumn{1}{c|}{\bf Error type} &
\multicolumn{1}{c|}{\bf Example sentence} & \multicolumn{1}{c|}{\bf Correction (edit)} \\ \hline
ArtOrDet & Article or determiner & In {\it late} nineteenth century, there was a severe air crash happening at
Miami international airport. & late $\rightarrow$ the late \\ \hline
Prep & Preposition & Also tracking people is very dangerous if it has
been controlled by bad men {\it in} a not good purpose. & in $\rightarrow$ for \\ \hline
Nn & Noun number & I think such powerful {\it device} shall not be
made easily available. & device $\rightarrow$ devices \\ \hline
Vform & Verb form & However, it is an achievement as it is an indication that our society
is {\it progressed} well and people are living in better conditions. &
progressed $\rightarrow$ progressing \\ \hline
SVA & Subject-verb agreement & People still {\it prefers} to bear the risk and allow their pets to have
maximum freedom. & prefers $\rightarrow$ prefer \\ \hline
\end{tabularx}
\caption{\label{tab:error-types} The five error types in our shared task.  }
\end{center}
\end{table*}

Since there are five error types in our shared task compared to two in
HOO 2012, there is a greater chance of encountering multiple,
interacting errors in a sentence in our shared task. This increases
the complexity of our shared task relative to that of HOO 2012. To
illustrate, consider the following sentence:

\begin{quote}
Although we have to admit some bad {\it effect} which {\it is} brought
by the new technology, still the advantages of the new technologies
cannot be simply discarded.
\end{quote}

\noindent The noun number error {\it effect} needs to be corrected
(effect $\rightarrow$ effects). This necessitates the correction of a
subject-verb agreement error (is $\rightarrow$ are). A pipeline system
in which corrections for subject-verb agreement errors occur strictly
before corrections for noun number errors would not be able to arrive
at a fully corrected sentence for this example. The ability to correct
multiple, interacting errors is thus necessary in our shared task. The
recent work of \cite{dahlmeier_emnlp12}, for example, is designed to
deal with multiple, interacting errors.

Note that the essays in the training data and the test essays
naturally contain grammatical errors of all types, beyond the five
error types focused in our shared task. In the automatically corrected
essays returned by a participating system, only corrections necessary
to correct errors of the five types are made. The other errors are to
be left uncorrected.

\section{Data}
\label{sec:data}

This section describes the training and test data released to each
participating team in our shared task.

\subsection{Training Data}
\label{sec:data_training}

The training data provided in our shared task is the NUCLE corpus, the
NUS Corpus of Learner English~\cite{dahlmeier_building_2013}. As noted
by \cite{leacock2010automated}, the lack of a manually annotated and
corrected corpus of English learner texts has been an impediment to
progress in grammatical error correction, since it prevents
comparative evaluations on a common benchmark test data set. NUCLE was
created precisely to fill this void. It is a collection of 1,414 essays
written by students at the National University of Singapore (NUS) who
are non-native speakers of English. The essays were written in
response to some prompts, and they cover a wide range of topics, such
as environmental pollution, health care, etc.  The grammatical errors
in these essays have been hand-corrected by professional English
instructors at NUS. For each grammatical error instance, the start and
end character offsets of the erroneous text span are marked, and the
error type and the correction string are provided. Manual annotation
is carried out using a graphical user interface specifically built for
this purpose.  The error annotations are saved as stand-off
annotations, in SGML format.

To illustrate, consider the following sentence at the start of the
first paragraph of an essay:

\begin{quote}
From past to the present, many important innovations have surfaced.
\end{quote}

\noindent There is an article/determiner error (past $\rightarrow$ the
past) in this sentence. The error annotation, also called {\it
correction} or {\it edit}, in SGML format is shown in
Figure~\ref{fig:error-annotation}.  {\tt start\_par} ({\tt end\_par})
denotes the paragraph ID of the start (end) of the erroneous text span
(paragraph ID starts from 0 by convention). {\tt start\_off} ({\tt
end\_off}) denotes the character offset of the start (end) of the
erroneous text span (again, character offset starts from 0 by
convention). The error tag is {\tt ArtOrDet}, and the correction
string is {\tt the past}.

\begin{figure*}[t]
{\tt
<MISTAKE start\_par="0" start\_off="5" end\_par="0" end\_off="9"> \\
<TYPE>ArtOrDet</TYPE> \\
<CORRECTION>the past</CORRECTION> \\
</MISTAKE>
}
\caption{An example error annotation.}
\label{fig:error-annotation}
\end{figure*}

The NUCLE corpus was first used in \cite{dahlmeier_acl11}, and has
been publicly available for research purposes since June
2011\footnote{http://www.comp.nus.edu.sg/$\sim$nlp/corpora.html}.  All
instances of grammatical errors are annotated in NUCLE, and the errors
are classified into 27 error types \cite{dahlmeier_building_2013}.

To help participating teams in their preparation for the shared task,
we also performed automatic preprocessing of the NUCLE corpus and
released the preprocessed form of NUCLE. The preprocessing operations
performed on the NUCLE essays include sentence segmentation and word
tokenization using the NLTK toolkit \cite{nltk09}, and part-of-speech
(POS) tagging, constituency and dependency tree parsing using the
Stanford parser \cite{klein03,manning06}. The error annotations, which
are originally at the character level, are then mapped to error
annotations at the word token level. Error annotations at the word
token level also facilitate scoring, as we will see in
Section~\ref{sec:eval}, since our scorer operates by matching
tokens. Note that although we released our own preprocessed version of
NUCLE, the participating teams were however free to perform their own
preprocessing if they so preferred.

\subsubsection{Revised version of NUCLE}

NUCLE release version 2.3 was used in the CoNLL-2013 shared task. In
this version, 17 essays were removed from the first release of NUCLE
since these essays were duplicates with multiple annotations. 

In the original NUCLE corpus, there is not an explicit preposition
error type. Instead, preposition errors are part of the Wcip (wrong
collocation/idiom/preposition) and Rloc (local redundancy) error
types. The Wcip error type combines errors concerning collocations,
idioms, and prepositions together into one error type.  The Rloc error
type annotates extraneous words which are redundant and should be
removed, and they include redundant articles, determiners, and
prepositions. In our shared task, in order to facilitate the detection
and correction of article/determiner errors and preposition errors, we
performed automatic mapping of error types in the original NUCLE
corpus. The mapping relies on POS tags, constituent parse trees, and
error annotations at the word token level.  Specifically, we map the
error types Wcip and Rloc to Prep, Wci, ArtOrDet, and
Rloc$-$. Prepositions in the error type Wcip or Rloc are mapped to a
new error type Prep, and redundant articles or determiners in the
error type Rloc are mapped to ArtOrDet.  The remaining unaffected Wcip
errors are assigned the new error type Wci and the remaining
unaffected Rloc errors are assigned the new error type Rloc$-$. The
code that performs automatic error type mapping was also provided to
the participating teams.

The statistics of the NUCLE corpus (release 2.3 version) are shown in
Table~\ref{tab:data-stats}. The distribution of errors among the five
error types is shown in Table~\ref{tab:error-dist}. The newly added
noun number error type in our shared task accounts for the second
highest number of errors among the five error types. The five error
types in our shared task constitute 35\% of all grammatical errors in
the training data, and 47\% of all errors in the test data. These
figures support our choice of these five error types to be the focus
of our shared task, since they account for a large percentage of all
grammatical errors in English learner essays.

While the NUCLE corpus is provided in our shared task, participating
teams are free to not use NUCLE, or to use additional resources and
tools in building their grammatical error correction systems, as long
as these resources and tools are publicly available and not
proprietary. For example, participating teams are free to use the
Cambridge FCE corpus \cite{briscoe11,nicholls03} (the training data
provided in HOO 2012 \cite{dale_hoo_2012}) as additional training
data.

\begin{table}
\begin{center}
\begin{tabular}{|l|r|r|} \hline
& \multicolumn{1}{c|}{\bf Training data} & \multicolumn{1}{c|}{\bf Test data} \\
& \multicolumn{1}{c|}{\bf (NUCLE)} & \\ \hline
\# essays      &     1,397 &     50 \\ \hline
\# sentences   &    57,151 &  1,381 \\ \hline
\# word tokens & 1,161,567 & 29,207 \\ \hline
\end{tabular}
\caption{\label{tab:data-stats} Statistics of training and test data.}
\end{center}
\end{table}

\begin{table}
\begin{center}
\begin{tabular}{|l|r|r|r|r|} \hline
\multicolumn{1}{|c|}{\bf Error tag} & \multicolumn{1}{c|}{\bf Training} & \multicolumn{1}{c|}{\bf \%} 
& \multicolumn{1}{c|}{\bf Test} & \multicolumn{1}{c|}{\bf \%} \\
& \multicolumn{1}{c|}{\bf data} & & \multicolumn{1}{c|}{\bf data} & \\
& \multicolumn{1}{c|}{\bf (NUCLE)} & & & \\ \hline
ArtOrDet  &  6,658 &  14.8 &   690 &  19.9 \\ \hline
Prep      &  2,404 &   5.3 &   312 &   9.0 \\ \hline
Nn        &  3,779 &   8.4 &   396 &  11.4 \\ \hline
Vform     &  1,453 &   3.2 &   122 &   3.5 \\ \hline
SVA       &  1,527 &   3.4 &   124 &   3.6 \\ \hline
5 types   & 15,821 &  35.1 & 1,644 &  47.4 \\ \hline
all types & 45,106 & 100.0 & 3,470 & 100.0 \\ \hline
\end{tabular}
\caption{\label{tab:error-dist} Error type distribution of the
training and test data.}
\end{center}
\end{table}

\subsection{Test Data}
\label{sec:data_test}

25 NUS students, who are non-native speakers of English, 
were recruited to write new essays to be used as blind test data in
the shared task. Each student wrote two essays in response to the two
prompts shown in Table~\ref{tab:prompts}, one essay per prompt. 
Essays written using the first prompt are present in the NUCLE
training data, while the second prompt is a new prompt not used
previously. As a result, 50 test essays were collected. The statistics
of the test essays are shown in Table~\ref{tab:data-stats}.

Error annotation on the test essays was carried out by a native
speaker of English who is a lecturer at the NUS Centre for English
Language Communication. The distribution of errors in the test essays
among the five error types is shown in Table~\ref{tab:error-dist}.
The test essays were then preprocessed in the same manner as the NUCLE
corpus. The preprocessed test essays were released to the
participating teams.

Unlike the test data used in HOO 2012 which was proprietary and not
available after the shared task, the test essays and their error
annotations in the CoNLL-2013 shared task are freely available after
the shared task.

\begin{table*}
\begin{center}
\begin{tabularx}{\textwidth}{|c|X|} \hline
\multicolumn{1}{|c|}{\bf ID} & \multicolumn{1}{c|}{\bf Prompt} \\ \hline
1 & Surveillance technology such as RFID (radio-frequency identification)
should not be used to track people (e.g., human implants and RFID tags
on people or products). Do you agree? Support your argument with
concrete examples. \\ \hline
2 & Population aging is a global phenomenon. Studies have shown that the
current average life span is over 65. Projections of the United
Nations indicate that the population aged 60 or over in developed and
developing countries is increasing at 2\% to 3\% annually. Explain why
rising life expectancies can be considered both a challenge and an
achievement. \\ \hline
\end{tabularx}
\caption{\label{tab:prompts} The two prompts used for the test essays.}
\end{center}
\end{table*}

\section{Evaluation Metric and Scorer}
\label{sec:eval}

A grammatical error correction system is evaluated by how well its
proposed corrections or edits match the gold-standard edits.  An essay
is first sentence-segmented and tokenized before evaluation is carried
out on the essay. To illustrate, consider the following tokenized
sentence $S$ written by an English learner:

\begin{quote}
There is no {\bf a doubt}, tracking {\bf system has} brought many
benefits in this information age .
\end{quote}

\noindent The set of gold-standard edits of a human annotator is
$\mathbf{g}$ = \{a doubt $\rightarrow$ doubt, system $\rightarrow$
systems, has $\rightarrow$ have\}. Suppose the tokenized output
sentence $H$ of a grammatical error correction system given the above
sentence is:

\begin{quote}
There is no doubt, tracking system has brought many
benefits in this information age .
\end{quote}

That is, the set of system edits is $\mathbf{e}$ = \{a doubt
$\rightarrow$ doubt\}. The performance of the grammatical error
correction system is measured by how well the two sets $\mathbf{g}$ and
$\mathbf{e}$ match, in the form of recall $R$, precision $P$, and $F_1$
measure: $R = 1/3, P = 1/1, F_1 = 2RP/(R+P) = 1/2$.

More generally, given a set of $n$ sentences, where $\mathbf{g}_i$ is
the set of gold-standard edits for sentence $i$, and $\mathbf{e}_i$ is
the set of system edits for sentence $i$, recall, precision, and $F_1$
are defined as follows:
\begin{equation} \label{eq:recall}
  R=\frac{\sum_{i=1}^n{|\mathbf{g}_i \cap \mathbf{e}_i|}}{\sum_{i=1}^n{|\mathbf{g}_i|}}
\end{equation}
\begin{equation} \label{eq:prec}
  P=\frac{\sum_{i=1}^n{|\mathbf{g}_i \cap \mathbf{e}_i|}}{\sum_{i=1}^n{|\mathbf{e}_i|}}
\end{equation}
\begin{equation} \label{eq:f1}
  F_1=\frac{2 \times R \times P}{R + P}
\end{equation}
where the intersection between $\mathbf{g}_i$ and $\mathbf{e}_i$ for sentence $i$ is defined as
\begin{equation}
  \mathbf{g}_i \cap \mathbf{e}_i=\{e \in \mathbf{e}_i | \exists{g \in \mathbf{g}_i, match(g, e)}\}
\end{equation}
Evaluation by the HOO scorer \cite{dale_hoo_2011} is based on
computing recall, precision, and $F_1$ measure as defined above.

Note that there are multiple ways to specify a set of gold-standard
edits that denote the same corrections. For example, in the above
learner-written sentence $S$, alternative but equivalent sets of
gold-standard edits are \{a $\rightarrow$ $\epsilon$, system
$\rightarrow$ systems, has $\rightarrow$ have\}, \{a $\rightarrow$
$\epsilon$, system has $\rightarrow$ systems have\}, etc. Given the
same learner-written sentence $S$ and the same system output sentence
$H$ shown above, one would expect a scorer to give the same 
$R, P, F_1$ scores regardless of which of the equivalent sets of
gold-standard edits is specified by an annotator.

However, this is not the case with the HOO scorer. This is because the
HOO scorer uses GNU {\tt wdiff}\footnote{http://www.gnu.org/s/wdiff/}
to extract the differences between the learner-written sentence $S$
and the system output sentence $H$ to form a set of system
edits. Since in general there are multiple ways to specify a set of
gold-standard edits that denote the same corrections, the set of
system edits computed by the HOO scorer may not match the set of
gold-standard edits specified, leading to erroneous scores. In the
above example, the set of system edits computed by the HOO scorer for
$S$ and $H$ is \{a $\rightarrow$ $\epsilon$\}. Given that the set of
gold-standard edits $\mathbf{g}$ is \{a doubt $\rightarrow$ doubt,
system $\rightarrow$ systems, has $\rightarrow$ have\}, the scores
computed by the HOO scorer are $R = P = F_1 = 0$, which are erroneous.

The \emph{MaxMatch} (M$^{2}$)
scorer\footnote{http://www.comp.nus.edu.sg/$\sim$nlp/software.html}
\cite{dahlmeier_better_2012} was designed to overcome this limitation
of the HOO scorer. The key idea is that the set of system edits
automatically computed and used in scoring should be the set that
maximally matches the set of gold-standard edits specified by the
annotator. The M$^{2}$ scorer uses an efficient algorithm to search
for such a set of system edits using an edit lattice. In the above
example, given $S$, $H$, and $\mathbf{g}$, the M$^{2}$ scorer is able
to come up with the best matching set of system edits $\mathbf{e}$ =
\{a doubt $\rightarrow$ doubt\}, thus giving the correct scores $R =
1/3, P = 1/1, F_1 = 1/2$.  We use the M$^{2}$ scorer in the CoNLL-2013
shared task.

The original M$^{2}$ scorer implemented in
\cite{dahlmeier_better_2012} assumes that there is one set of
gold-standard edits $\mathbf{g}_i$ for each sentence $i$. However, it
is often the case that multiple alternative corrections are acceptable
for a sentence. As we allow participating teams to submit alternative
sets of gold-standard edits for a sentence, we also extend the M$^{2}$
scorer to deal with multiple alternative sets of gold-standard edits.

Based on Equations~\ref{eq:recall} and~\ref{eq:prec},
Equation~\ref{eq:f1} can be re-expressed as:

\begin{equation}\label{eq:F1_modified}
  F_1=\frac{2\times \sum_{i=1}^n{|\mathbf{g}_i \cap \mathbf{e}_i|}}{\sum_{i=1}^n{(|\mathbf{g}_i|+|\mathbf{e}_i|)}}
\end{equation}

To deal with multiple alternative sets of gold-standard edits
$\mathbf{g}_i$ for a sentence $i$, the extended M$^{2}$ scorer chooses
the $\mathbf{g}_i$ that maximizes the cumulative $F_1$ score for
sentences $1, \ldots, i$. Ties are broken based on the following
criteria: first choose the $\mathbf{g}_i$ that maximizes the numerator
$\sum_{i=1}^n{|\mathbf{g}_i \cap \mathbf{e}_i|}$, then choose the
$\mathbf{g}_i$ that minimizes the denominator
$\sum_{i=1}^n{(|\mathbf{g}_i|+|\mathbf{e}_i|)}$, finally choose the
$\mathbf{g}_i$ that appears first in the list of alternatives.

\section{Approaches}
\label{sec:approaches}

54 teams registered to participate in the shared task, out of which 17
teams submitted the output of their grammatical error correction
systems by the deadline. These teams are listed in
Table~\ref{tab:teams}. Each team is assigned a 3 to 4-letter team
ID. In the remainder of this paper, we will use the assigned team ID
to refer to a participating team. Every team submitted a system
description paper (the only exception is the SJT2 team).

\begin{table*}[htpb]
\begin{center}
\begin{tabular}{|l|l|} \hline
\multicolumn{1}{|c|}{\bf Team ID} & \multicolumn{1}{c|}{\bf Affiliation} \\ \hline
CAMB & University of Cambridge \\ \hline
HIT  & Harbin Institute of Technology \\ \hline
IITB & Indian Institute of Technology, Bombay \\ \hline
KOR  & Korea University \\ \hline
NARA & Nara Institute of Science and Technology \\ \hline
NTHU & National Tsing Hua University \\ \hline
SAAR & Saarland University \\ \hline
SJT1 & Shanghai Jiao Tong University (Team \#1) \\ \hline
SJT2 & Shanghai Jiao Tong University (Team \#2) \\ \hline
STAN & Stanford University \\ \hline
STEL & Stellenbosch University \\ \hline
SZEG & University of Szeged \\ \hline
TILB & Tilburg University \\ \hline
TOR  & University of Toronto \\ \hline
UAB  & Universitat Autònoma de Barcelona \\ \hline
UIUC & University of Illinois at Urbana-Champaign \\ \hline
UMC  & University of Macau \\ \hline
\end{tabular}
\caption{\label{tab:teams} The list of 17 participating teams.}
\end{center}
\end{table*}

\begin{sidewaystable*}[htpb]
\small
\centering
\begin{tabularx}{\textwidth}{|l|L{1cm}|C{1cm}|X|L{7cm}|L{4cm}|}\hline
\multicolumn{1}{|c|}{\bf Team} & \multicolumn{1}{c|}{\bf Error} & \multicolumn{1}{c|}{\bf Approach} &
\multicolumn{1}{c|}{\bf Description of Approach} & \multicolumn{1}{c|}{\bf Linguistic Features} &
\multicolumn{1}{c|}{\bf External Resources} \\\hline
CAMB & ANPSV & T & factored phrase-based translation model with IRST language model & lexical, POS & Cambridge Learner Corpus \\\hline
HIT & ANP & M & maximum entropy with confidence tuning, and genetic algorithm for feature selection & lexical, POS, constituency parse, dependency parse, semantic & WordNet, Longman dictionary \\\cline{2-5}
& SV & R & rule-based & POS, dependency parse, semantic & \\\hline
IITB & AN & M & maximum entropy & lexical, POS, noun properties & Wiktionary \\\cline{2-5}
& S & R & rule-based & POS, constituency parse, dependency parse & \\\hline
KOR & ANP & M & maximum entropy & lexical, POS, head-modifier, dependency parse & (none) \\\hline
NARA & AP & T & phrase-based statistical machine translation & lexical & Lang-8 \\\cline{2-6}
& N & M & adaptive regularization of weight vectors & lexical, lemma, constituency parse & Gigaword \\\cline{2-6}
& SV & L & treelet (tree-based) language model & lexical, POS, constituency parse & Penn Treebank, Gigaword \\\hline
NTHU & ANPV & L & n-gram-based and dependency-based language model & lexical, POS, constituency parse, dependency parse & Google Web-1T \\\hline
SAAR & A & M & multi-class SVM and naive Bayes & lexical, POS, constituency parse & CMU Pronouncing Dictionary \\\cline{2-5}
& S & R & rule-based & POS, dependency parse & \\\hline
SJT1 & ANPSV & M & maximum entropy (with LM post-filtering) & lexical, lemma, POS, constituency parse, dependency parse & Europarl \\\hline
STAN & ANPSV & H & English Resource Grammar (ERG), head-driven phrase structure, extended with hand-coded mal-rules & lexical, POS, constituency parse, semantic & English Resource Grammar \\\hline
STEL & ANPSV & T & tree-to-string with GHKM transducer & constituency parse & Wikipedia, WordNet \\\hline
SZEG & AN & M & maximum entropy & LFG, lexical, constituency parse, dependency parse & (none) \\\hline
TILB & ANPSV & M & binary and multi-class IGTree & lexical, lemma, POS & Google Web-1T, Gigaword \\\hline
TOR & ANPSV & T & noisy channel model involving transformation of single words & lexical, POS & Wikipedia \\\hline
UAB & ANPSV & R & rule-based & lexical, dependency parse & Top 250 uncountable nouns, FreeLing morphological dictionary \\\hline
UIUC & ANPSV & M & A: multi-class averaged perceptron; others: naive Bayes & lexical, POS, shallow parse & Google Web-1T, Gigaword \\\hline
UMC & ANPSV & H & pipeline: rule-based filter $\rightarrow$ semi-supervised multi-class maximum entropy classifier $\rightarrow$ LM confidence scorer & lexical, POS, dependency parse & News corpus, JMySpell dictionary,  Google Web-1T, Penn Treebank \\\hline
\end{tabularx}
\caption{\label{tab:team-profile} Profile of the participating
teams. The {\it Error} column shows the error type, where each letter
denotes the error type beginning with that initial letter. The {\it
Approach} column shows the approach adopted by each team, sometimes
broken down according to the error type: H denotes a hybrid classifier
approach, L denotes a language modeling-based approach, M denotes a
machine learning-based classifier approach, R denotes a rule-based
classifier (non-machine learning) approach, and T denotes a machine
translation approach}
\end{sidewaystable*}

Many different approaches are adopted by participating teams in the
CoNLL-2013 shared task, and Table~\ref{tab:team-profile} summarizes
these approaches.  A commonly used approach in the shared task and in
grammatical error correction research in general is to build a
classifier for each error type. For example, the classifier for noun
number returns the classes \{singular, plural\}, the classifier for
article returns the classes \{a/an, the, $\epsilon$\}, etc. The
classifier for an error type may be learned from training examples
encoding the surrounding context of an error occurrence, or may be
specified by deterministic hand-crafted rules, or may be built using a
hybrid approach combining both machine learning and hand-crafted
rules. These approaches are denoted by M, R, and H respectively in
Table~\ref{tab:team-profile}.

The machine translation approach (denoted by T in
Table~\ref{tab:team-profile}) to grammatical error correction treats
the task as ``translation'' from bad English to good English. Both
phrase-based translation and syntax-based translation approaches are
used by teams in the CoNLL-2013 shared task. Another related approach
is the language modeling approach (denoted by L in
Table~\ref{tab:team-profile}), in which the probability of a learner
sentence is compared with the probability of a candidate corrected
sentence, based on a language model built from a background
corpus. The candidate correction is chosen if it results in a
corrected sentence with a higher probability. In general, these
approaches are not mutually exclusive. For example, the work of
\cite{dahlmeier_emnlp12,nara-conll13} includes elements of machine
learning-based classification, machine translation, and language
modeling approaches.

When different approaches are used to tackle different error types by
a system, we break down the error types into different rows in
Table~\ref{tab:team-profile}, and specify the approach used for each
group of error types. For instance, the HIT team uses a machine
learning approach to deal with article/determiner, noun number, and
preposition errors, and a rule-based approach to deal with
subject-verb agreement and verb form errors. As such, the entry for
HIT is sub-divided into two rows, to make it clear which particular
error type is handled by which approach.

Table~\ref{tab:team-profile} also shows the linguistic features used
by the participating teams, which include lexical features (i.e.,
words, collocations, n-grams), parts-of-speech (POS), constituency
parses, dependency parses, and semantic features (including semantic role
labels).

While all teams in the shared task use the NUCLE corpus, they are also
allowed to use additional external resources (both corpora and tools)
so long as they are publicly available and not proprietary. The
external resources used by the teams are also listed in
Table~\ref{tab:team-profile}.

\section{Results}
\label{sec:results}

\begin{table}[htbp]
\begin{center}
\begin{tabular}{|c|l|r|r|r|}\hline
\multicolumn{1}{|c|}{\bf Rank} & \multicolumn{1}{c|}{\bf Team} & 
\multicolumn{1}{c|}{$\mathbf{R}$} & \multicolumn{1}{c|}{$\mathbf{P}$} &
\multicolumn{1}{c|}{$\mathbf{F}_1$} \\ \hline
1 & UIUC & 23.49 & 46.45 &  31.20 \\ \hline
2 & NTHU & 26.35 & 23.80 &  25.01 \\ \hline
3 & HIT & 16.56 & 35.65 &  22.61 \\ \hline
4 & NARA & 18.62 & 27.39 &  22.17 \\ \hline
5 & UMC & 17.53 & 28.49 &  21.70 \\ \hline
6 & STEL & 13.33 & 27.00 &  17.85 \\ \hline
7 & SJT1 & 10.96 & 40.18 &  17.22 \\ \hline
8 & CAMB & 10.10 & 39.15 &  16.06 \\ \hline
9 & IITB & 4.99 & 28.18 &  8.48 \\ \hline
10 & STAN & 4.69 & 25.50 &  7.92 \\ \hline
11 & TOR & 4.81 & 17.67 &  7.56 \\ \hline
12 & KOR & 3.71 & 43.88 &  6.85 \\ \hline
13 & TILB & 7.24 & 6.25 &  6.71 \\ \hline
14 & SZEG & 3.16 & 5.52 &  4.02 \\ \hline
15 & UAB & 1.22 & 12.42 &  2.22 \\ \hline
16 & SAAR & 1.10 & 27.69 &  2.11 \\ \hline
17 & SJT2 & 0.24 & 13.33 &  0.48 \\ \hline
\end{tabular}
\end{center}
\caption{\label{tab:score-noalt} Scores (in \%) without alternative answers.}
\end{table}

All submitted system output was evaluated using the M$^{2}$ scorer,
based on the error annotations provided by our annotator. The recall
($R$), precision ($P$), and $F_1$ measure of all teams are shown in
Table~\ref{tab:score-noalt}. The performance of the teams varies
greatly, from barely half a per cent to 31.20\% for the top team.

The nature of grammatical error correction is such that multiple,
different corrections are often acceptable.  In order to allow the
participating teams to raise their disagreement with the original
gold-standard annotations provided by the annotator, and not
understate the performance of the teams, we allow the teams to submit
their proposed alternative answers. This was also the practice adopted
in HOO 2011 and HOO 2012. Specifically, after the teams submitted
their system output and the error annotations on the test essays were
released, we allowed the teams to propose alternative answers
(gold-standard edits), to be submitted within four days after the
initial error annotations were released.  The same annotator who
provided the error annotations on the test essays also judged the
alternative answers proposed by the teams, to ensure consistency. In
all, five teams (NTHU, STEL, TOR, UIUC, UMC) submitted alternative
answers.

The same submitted system output was then evaluated using the extended
M$^{2}$ scorer, with the original annotations augmented with the
alternative answers.  Table~\ref{tab:score-alt} shows the recall
($R$), precision ($P$), and $F_1$ measure of all teams under this new
evaluation setting.

The $F_1$ measure of every team improves when evaluated with
alternative answers. Not surprisingly, the teams which submitted
alternative answers tend to show the greatest improvements in their
$F_1$ measure. Overall, the UIUC team \cite{uiuc-conll13} achieves the
best $F_1$ measure, with a clear lead over the other teams in the
shared task, under both evaluation settings (without and with
alternative answers).

For future research which uses the test data of the CoNLL-2013 shared
task, we recommend that evaluation be carried out in the setting that
does {\it not} use alternative answers, to ensure a fairer evaluation.
This is because the scores of the teams which submitted alternative
answers tend to be higher in a biased way when evaluated with
alternative answers.

\begin{table}[htbp]
\begin{center}
\begin{tabular}{|c|l|r|r|r|}\hline
\multicolumn{1}{|c|}{\bf Rank} & \multicolumn{1}{c|}{\bf Team} & 
\multicolumn{1}{c|}{$\mathbf{R}$} & \multicolumn{1}{c|}{$\mathbf{P}$} &
\multicolumn{1}{c|}{$\mathbf{F}_1$} \\ \hline
1 & UIUC & 31.87 & 62.19 &  42.14 \\ \hline
2 & NTHU & 34.62 & 30.57 &  32.46 \\ \hline
3 & UMC & 23.66 & 37.12 &  28.90 \\ \hline
4 & NARA & 24.05 & 33.92 &  28.14 \\ \hline
5 & HIT & 20.29 & 41.75 &  27.31 \\ \hline
6 & STEL & 18.91 & 37.12 &  25.05 \\ \hline
7 & CAMB & 14.19 & 52.11 &  22.30 \\ \hline
8 & SJT1 & 13.67 & 47.77 &  21.25 \\ \hline
9 & TOR & 8.77 & 30.67 &  13.64 \\ \hline
10 & IITB & 6.55 & 34.93 &  11.03 \\ \hline
11 & STAN & 5.86 & 29.93 &  9.81 \\ \hline
12 & KOR & 4.78 & 53.24 &  8.77 \\ \hline
13 & TILB & 9.29 & 7.60 &  8.36 \\ \hline
14 & SZEG & 4.07 & 6.67 &  5.06 \\ \hline
15 & UAB & 1.81 & 17.39 &  3.28 \\ \hline
16 & SAAR & 1.68 & 40.00 &  3.23 \\ \hline
17 & SJT2 & 0.33 & 16.67 &  0.64 \\ \hline
\end{tabular}
\end{center}
\caption{\label{tab:score-alt} Scores (in \%) with alternative answers.}
\end{table}

\begin{sidewaystable*}[htbp]
\begin{center}
\begin{tabular}{|l|r|r|r|r|r|r|r|r|r|r|r|r|}\hline
\multicolumn{1}{|c|}{\multirow{2}{*}{\bf Team}} & \multicolumn{3}{c|}{\bf ArtOrDet} & \multicolumn{3}{c|}{\bf Prep} & \multicolumn{3}{c|}{\bf Nn} & \multicolumn{3}{c|}{\bf Vform/SVA} \\ \cline{2-13}
 & \multicolumn{1}{c|}{$\mathbf{R}$} & \multicolumn{1}{c|}{$\mathbf{P}$} & \multicolumn{1}{c|}{$\mathbf{F}_1$} & \multicolumn{1}{c|}{$\mathbf{R}$} & \multicolumn{1}{c|}{$\mathbf{P}$} & \multicolumn{1}{c|}{$\mathbf{F}_1$} & \multicolumn{1}{c|}{$\mathbf{R}$} & \multicolumn{1}{c|}{$\mathbf{P}$} & \multicolumn{1}{c|}{$\mathbf{F}_1$} & \multicolumn{1}{c|}{$\mathbf{R}$} & \multicolumn{1}{c|}{$\mathbf{P}$} & \multicolumn{1}{c|}{$\mathbf{F}_1$} \\ \hline
CAMB & 15.07 & 38.66 & 21.69 & 3.54 & 40.74 & 6.51 & 7.58 & 55.56 & 13.33 & 8.54 & 31.82 & 13.46\\\hline
HIT & 24.20 & 42.82 & 30.93 & 2.89 & 28.12 & 5.25 & 17.17 & 29.69 & 21.76 & 11.38 & 26.42 & 15.91\\\hline
IITB & 1.30 & 21.43 & 2.46 & \multicolumn{3}{c|}{(not done)} & 9.85 & 28.68 & 14.66 & 13.82 & 30.09 & 18.94\\\hline
KOR & 4.78 & 53.23 & 8.78 & 0.32 & 4.76 & 0.60 & 6.82 & 49.09 & 11.97 & \multicolumn{3}{c|}{(not done)}\\\hline
NARA & 20.43 & 34.06 & 25.54 & 12.54 & 29.10 & 17.53 & 16.41 & 48.87 & 24.57 & 24.80 & 14.81 & 18.54\\\hline
NTHU & 21.01 & 35.80 & 26.48 & 12.86 & 12.01 & 12.42 & 45.96 & 40.90 & 43.28 & 26.83 & 12.22 & 16.79\\\hline
SAAR & 0.72 & 62.50 & 1.43 & \multicolumn{3}{c|}{(not done)} & \multicolumn{3}{c|}{(not done)} & 5.28 & 23.21 & 8.61\\\hline
SJT1 & 16.81 & 47.15 & 24.79 & 1.29 & 12.50 & 2.33 & 13.64 & 42.19 & 20.61 & 2.44 & 14.63 & 4.18\\\hline
SJT2 & 0.00 & 0.00 & 0.00 & 0.00 & 0.00 & 0.00 & 1.01 & 13.33 & 1.88 & 0.00 & 0.00 & 0.00\\\hline
STAN & 3.91 & 20.45 & 6.57 & 0.32 & 20.00 & 0.63 & 6.06 & 29.63 & 10.06 & 10.16 & 32.05 & 15.43\\\hline
STEL & 12.61 & 27.71 & 17.33 & 9.32 & 25.66 & 13.68 & 18.18 & 46.75 & 26.18 & 12.60 & 17.61 & 14.69\\\hline
SZEG & 1.16 & 1.70 & 1.38 & \multicolumn{3}{c|}{(not done)} & 11.11 & 13.62 & 12.24 & \multicolumn{3}{c|}{(not done)}\\\hline
TILB & 4.49 & 4.49 & 4.49 & 10.61 & 5.07 & 6.86 & 7.07 & 21.21 & 10.61 & 10.98 & 9.57 & 10.23\\\hline
TOR & 8.55 & 25.54 & 12.81 & 2.25 & 5.38 & 3.17 & 1.77 & 31.82 & 3.35 & 2.44 & 12.24 & 4.07\\\hline
UAB & 0.00 & 0.00 & 0.00 & 0.00 & 0.00 & 0.00 & 0.00 & 0.00 & 0.00 & 8.13 & 12.42 & 9.83\\\hline
UIUC & 25.65 & 47.84 & 33.40 & 4.18 & 26.53 & 7.22 & 38.38 & 52.23 & 44.25 & 17.89 & 38.94 & 24.51\\\hline
UMC & 21.01 & 30.27 & 24.81 & 1.93 & 35.29 & 3.66 & 23.23 & 27.96 & 25.38 & 18.29 & 28.85 & 22.39\\\hline
\end{tabular}
\end{center}
\caption{\label{tab:pertype-noalt} Scores (in \%) without alternative
answers, distinguished by error type. If a team indicates that its
system does not handle a particular error type, its entry for that
error type is marked as ``(not done)".}
\end{sidewaystable*}
\begin{sidewaystable*}[htbp]
\begin{center}
\begin{tabular}{|l|r|r|r|r|r|r|r|r|r|r|r|r|}\hline
\multicolumn{1}{|c|}{\multirow{2}{*}{\bf Team}} & \multicolumn{3}{c|}{\bf ArtOrDet} & \multicolumn{3}{c|}{\bf Prep} & \multicolumn{3}{c|}{\bf Nn} & \multicolumn{3}{c|}{\bf Vform/SVA} \\ \cline{2-13}
 & \multicolumn{1}{c|}{$\mathbf{R}$} & \multicolumn{1}{c|}{$\mathbf{P}$} & \multicolumn{1}{c|}{$\mathbf{F}_1$} & \multicolumn{1}{c|}{$\mathbf{R}$} & \multicolumn{1}{c|}{$\mathbf{P}$} & \multicolumn{1}{c|}{$\mathbf{F}_1$} & \multicolumn{1}{c|}{$\mathbf{R}$} & \multicolumn{1}{c|}{$\mathbf{P}$} & \multicolumn{1}{c|}{$\mathbf{F}_1$} & \multicolumn{1}{c|}{$\mathbf{R}$} & \multicolumn{1}{c|}{$\mathbf{P}$} & \multicolumn{1}{c|}{$\mathbf{F}_1$} \\ \hline
CAMB & 19.62 & 49.81 & 28.15 & 5.04 & 50.00 & 9.15 & 9.50 & 69.09 & 16.70 & 16.52 & 54.41 & 25.34\\\hline
HIT & 27.41 & 47.44 & 34.74 & 4.58 & 37.50 & 8.16 & 19.95 & 35.37 & 25.51 & 17.90 & 38.32 & 24.40\\\hline
IITB & 1.79 & 28.57 & 3.37 & \multicolumn{3}{c|}{(not done)} & 11.91 & 35.29 & 17.81 & 18.67 & 36.84 & 24.78\\\hline
KOR & 5.95 & 64.52 & 10.90 & 1.53 & 19.05 & 2.83 & 7.52 & 54.55 & 13.22 & \multicolumn{3}{c|}{(not done)}\\\hline
NARA & 25.79 & 43.34 & 32.34 & 17.60 & 34.56 & 23.33 & 19.15 & 57.04 & 28.68 & 34.62 & 19.10 & 24.62\\\hline
NTHU & 25.30 & 42.54 & 31.73 & 20.45 & 16.17 & 18.06 & 52.35 & 50.87 & 51.60 & 43.03 & 19.22 & 26.57\\\hline
SAAR & 1.04 & 87.50 & 2.06 & \multicolumn{3}{c|}{(not done)} & \multicolumn{3}{c|}{(not done)} & 8.60 & 33.93 & 13.72\\\hline
SJT1 & 19.65 & 54.07 & 28.82 & 1.92 & 15.62 & 3.42 & 16.46 & 52.34 & 25.05 & 4.05 & 21.95 & 6.84\\\hline
SJT2 & 0.00 & 0.00 & 0.00 & 0.00 & 0.00 & 0.00 & 1.26 & 16.67 & 2.35 & 0.00 & 0.00 & 0.00\\\hline
STAN & 4.91 & 24.81 & 8.20 & 0.38 & 20.00 & 0.75 & 6.78 & 33.33 & 11.27 & 13.51 & 37.97 & 19.93\\\hline
STEL & 16.23 & 35.69 & 22.31 & 12.83 & 29.82 & 17.94 & 26.05 & 70.00 & 37.97 & 20.52 & 26.55 & 23.15\\\hline
SZEG & 1.50 & 2.13 & 1.76 & \multicolumn{3}{c|}{(not done)} & 12.87 & 15.95 & 14.25 &  \multicolumn{3}{c|}{(not done)}\\\hline
TILB & 5.78 & 5.64 & 5.71 & 13.91 & 5.68 & 8.07 & 8.25 & 24.26 & 12.31 & 16.36 & 12.77 & 14.34\\\hline
TOR & 13.10 & 39.13 & 19.63 & 5.97 & 12.31 & 8.04 & 3.52 & 63.64 & 6.67 & 8.14 & 35.29 & 13.24\\\hline
UAB & 0.00 & 0.00 & 0.00 & 0.00 & 0.00 & 0.00 & 0.00 & 0.00 & 0.00 & 12.61 & 17.39 & 14.62\\\hline
UIUC & 31.99 & 59.84 & 41.69 & 8.81 & 46.94 & 14.84 & 46.88 & 70.00 & 56.15 & 28.57 & 60.71 & 38.86\\\hline
UMC & 25.88 & 36.74 & 30.37 & 3.47 & 56.25 & 6.55 & 30.61 & 38.64 & 34.16 & 26.81 & 40.13 & 32.14\\\hline
\end{tabular}
\end{center}
\caption{\label{tab:pertype-alt} Scores (in \%) with alternative
answers, distinguished by error type. If a team indicates that its
system does not handle a particular error type, its entry for that
error type is marked as ``(not done)".}
\end{sidewaystable*}

We are also interested in the analysis of scores of each of the five
error types. To compute the recall of an error type, we need to know
the error type of each gold-standard edit, which is provided by the
annotator. To compute the precision of each error type, we need to
know the error type of each system edit, which however is not
available since the submitted system output only contains the
corrected sentences with no indication of the error type of the system
edits.

In order to determine the error type of system edits, we first perform
POS tagging on the submitted system output using the Stanford parser
\cite{klein03}. We also make use of the POS tags assigned in
the preprocessed form of the test essays. We then assign an error type
to a system edit based on the automatically determined POS tags, as
follows:

\begin{itemize}

\item ArtOrDet: The system edit involves a change (insertion,
deletion, or substitution) of words tagged as article/determiner,
i.e., DT or PDT.

\item Prep: The system edit involves a change of words tagged as
preposition, i.e., IN or TO.

\item Nn: The system edit involves a change of words such that a word
in the source string is a singular noun (tagged as NN or NNP) and a
word in the replacement string is a plural noun (tagged as NNS or
NNPS), or vice versa. Since a word tagged as JJ (adjective) can serve
as a noun, a system edit that involves a change of POS tags from JJ to
one of \{NN, NNP, NNS, NNPS\} or vice versa also qualifies.

\item Vform/SVA: The system edit involves a change of words tagged as
one of the verb POS tags, i.e., VB, VBD, VBG, VBN, VBP, and VBZ.

\end{itemize}

The verb form and subject-verb agreement error types are grouped
together into one category, since it is difficult to automatically
distinguish the two in a reliable way.

The scores when distinguished by error type are shown in
Tables~\ref{tab:pertype-noalt} and~\ref{tab:pertype-alt}. Based on the
$F_1$ measure of each error type, the noun number error type gives the
highest scores, and preposition errors remain the most challenging
error type to correct.

\section{Conclusions}
\label{sec:conclusions}

The CoNLL-2013 shared task saw the participation of 17 teams worldwide
to evaluate their grammatical error correction systems on a common
test set, using a common evaluation metric and scorer. The five error
types included in the shared task account for at least one-third to
close to one-half of all errors in English learners' essays. The best
system in the shared task achieves an $F_1$ score of 42\%, when it is
scored with multiple acceptable answers. There is still much room for
improvement, both in the accuracy of grammatical error correction
systems, and in the coverage of systems to deal with a more
comprehensive set of error types.  The evaluation data sets and scorer
used in our shared task serve as a benchmark for future research on
grammatical error
correction\footnote{http://www.comp.nus.edu.sg/$\sim$nlp/conll13st.html}.

\section*{Acknowledgments}

This research is supported by the Singapore National Research
Foundation under its International Research Centre @ Singapore Funding
Initiative and administered by the IDM Programme Office.

\bibliographystyle{acl}
\bibliography{conll2013}

\end{document}